# A Network for structural dense displacement based on 3D deformable mesh model and optical flow

Peimian Du[1*], Qicheng Guo [2], Yanru Li [3], Yang Yu[4]
[1] *School of Civil Engineering, Harbin Institute of Technology, Harbin, China*
[2] *School of Civil Engineering, Harbin Institute of Technology, Harbin, China*
[3] *School of Civil Engineering, Harbin Institute of Technology, Harbin, China*
[4] *School of Civil Engineering, Harbin Institute of Technology, Harbin, China*

**ABSTRACT**

This study proposes a Network to recognize displacement of a RC frame structure from a video by a monocular camera. The proposed Network consists of two modules which is FlowNet2 and POFRN-Net. FlowNet2 is used to generate dense optical flow as well as POFRN-Net is to extract pose parameter *H*. FlowNet2 convert two video frames into dense optical flow. POFRN-Net is inputted dense optical flow from FlowNet2 to output the pose parameter *H*. The displacement of any points of structure can be calculated from parameter *H*. The Fast Fourier Transform (FFT) is applied to obtain frequency domain signal from corresponding displacement signal. Furthermore, the comparison of the truth displacement on the First floor of the First video is shown in this study. Finally, the predicted displacements on four floors of RC frame structure of given three videos are exhibited in the last of this study.

## 1   INTRODUCTION

Structural health monitoring (SHM) can simulate human self-sensing and self-diagnosis ability through sensing technology, and achieve the real-time perception of structural safety, structural performance state and structural performance evolution [1]. The SHM is considered as an inverse problem, which the structural defects will be detected based on the response of the structure under the excitation. SHM system consists of several parts, including various sensors, data-acquisition devices, data transmission system, database for data management, data analysis and modeling, condition assessment and performance prediction, alarm devices, visualization user interface, and software and operating system [2].

In the field of civil engineering, SHM is widely applied due to its ability to respond to adverse structural changes, improving structural reliability and life cycle management. However, the SHM system of complex large-scale structure require a huge number of traditional sensors and a high-density placement of sensors to obtain the accurate dynamic information of the structure, which cannot satisfy

*\*Corresponding Author:* Peimian Du
*Email:*  peimiandu@163.com

the cost limitation and information processing capability requirements [3].

In response to the defect, the Computer Vision technics was applied. The Computer Vision make it possible to observe the structural change from a holistic perspective. Besides, compare with the traditional methods, the contactless vision-based methods require less cost and less space.

This project aims to explore more effective data analysis methods and provide new ideas for students and young researchers in different fields to develop structural health monitoring technology. In this project, a four-story RC frame was constructed as the target object of the dynamic experiment and was placed on a shaking table. As presented in the figure 1, comprehensive measurement method was applied to gain the absolute acceleration and relative displacement of each story, including accelerometers, wire-type displacement sensors, and NDI Optotrack Certus high-speed motion capture system [4]. Besides, during the experiment, the response of the frame was recorded by a SONY HDR-PJ220 camera [4]. The given data including three videos of the experiment in different excitations, the displacement of the ground truth, and the acceleration data in the experiment. And the task is to determine the displacement of the frame and the modal in both frequency domain and time-frequency domain.

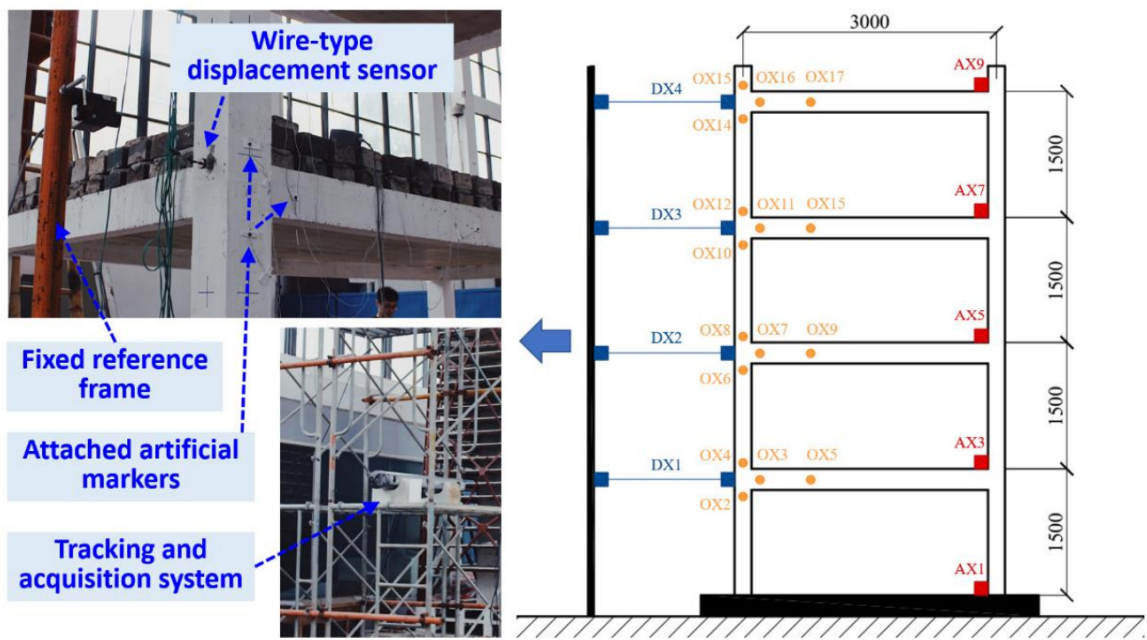

Fig. 1. Comprehensive Measurement Methods [4]

## 2　LITERATURE REVIEW

### 2.1　Structural Health Monitoring

Initially, the SHM system was employed to detect the damage in the aerospace engineering and industry. However, in the late 1970s, SHM was first used in civil engineering field. Farrar and Worden applied the on offshore platforms to investigate [5]. And in 1990s, Beck and Katafyglotis provide a probabilistic identification approach to monitor the changes in stiffness distribution of the structure component [6]. Mita investigate the rapidly emerging SHM in field of civil engineering in Japan [7]. In

1997, the earliest study on the health monitoring of linear systems was proposed by Al-Khalidy [8].

The emergence in sensing technologies and in computer technologies expand the application of the SHM system. In the past two decades, the SHM was considered as a human nervous system, where the processing unit will receive the data from sensors and do the judgment [9].

In 1993, Rytter provided a rating criterion to categorize the SHM according to the information obtained from the structure [10]. Level 1 is identification, where the existence of a defect on a global scale can be determined by the SHM. Level 2 is localization, where the location of the damage can be detected by SHM. Level 3 is assessment, where the intensity of damage in various components can be determined by the SHM. Level 4 is lifetime prediction, where the structure's remaining life can be estimated by the SHM.

Gharehbaghi suggested that the SHM system can be classified into 8 types due to the methods applied, traditional inspection, anomaly detection methods, reliability- based methods, acoustic emission (AE), feature-based, time-domain, computer vision, data-driven based machine learning [11].

The breakthrough in the computer vision technology and image acquisition equipment promotes the SHM based on the computer vision and data-driven based machine learning. Due to the advantage including long-distance, non-contact, high-precision, time-saving and labor-saving, and multi-point monitoring, the computer vision was wildly studied and applied [12]. Smarsly suggested that the data-driven based machine learning approaches can reduce the cost in computational operations and determine the unknown physical properties of the structure when enough sensors are available [13].

## 2.2 Computer Vision

Computer vision is an interdisciplinary technology that aims to derive information from the images or videos by simulating the human vision [14]. According to the discussion before, the computer vision was wildly studied and applied in the SHM in civil engineering field. This part will focus on the others work on the computer vision based SHM.

In early stage, Furquharson monitored the structure deformation of the Tacoma Narrows Bridge by the image data in 1954 [15]. Marécos monitored the change of the Tagus River Suspension Bridge during construction, load test and long-term observation up to 1971 by the multiple method including image [16]. The computer provides sufficient computation ability to analysis the embed information in the images and videos. Subsequently, the vision-based methods developed into two groups.

Processing-based approaches extract the features in the acquired images by processing the images and apply machine learning classification methods. Abdel-Qader provided edge detection filters to detect the edge of the object on the bridge in 2003 [17]. Choi created morphological features approach to morphological analyze and classify the surface corrosion damage in 2005 [18]. Lyasheva proposed bottom-hat transform approach to detect and recognize the crack on the pavement in 2019 [19]. Shan et al. applied the Canny-Zernike algorithm method to measure the width of the crack [20]. Qiang et al. provided an adaptive canny edge detection algorithm based on the Gauss filter and Otsu method to identify the crack [21]. Besides, the processing-based approach are also applied on the displacement recognition. Chen et al. used data from the high-speed cameras to determine the mode-shaped curvature of a beam [22]. Horn and Schunck proffered the classical variational optical flow estimation model, which is the starting point of optical flow methods [23]. Sarraf et al. carried out the perform operational analysis on a wind turbine blade by the base-based motion estimation to extract the resonance frequency and the defects [24]. The result shows that the base-based motion estimation could mitigate impact of noisy environment [24]. The study of Cabo et al. presents that PME had a good performance compare with the traditional sensing methods [25].

Deep learning-based approaches provide higher ability compare with the processing-based approaches due to the different detection approaches [26]. German et al. provided a damage index for concrete columns based on computer vision to quantify the damage rapidly after the earthquake [27]. Zhang et al. introduced a CNN called Crack Net to classify different cracks on asphalt surfaces [28]. Yang et al. promoted a Fast R-CNN to learn temperature gradient on the cracking surface of the steel plate generated by thermal imaging and a rolling electric heating rod [29]. Chen introduced a rotation-invariant fully convolutional network (FCN) called ARF-Crack, which can detect the crack on the pixel level by the application of FCN named DeepCrack and encoding the rotation-invariant characteristic into the network [30].

Jin et al. combined the processing-based approaches and deep learning-based approaches to identify the displacement and three-dimensional poses of structures [31]. The study used a U-net structure to predict the semantic classification masks through the RED function. And the obtained semantic classification masks were recognized as training set to train another CNN called Inception-Net, in which the output is the displacement and three-dimensional poses of structures. The result shows that, the combination of processing-based approaches and deep learning-based approaches improve efficiency while also improving accuracy.

## 3 METHODOLOGY

### 3.1 Overview of Methodology

The study proposes a network for a RC frame structure displacement identification from videos filmed by a monocular SONY HDR-PJ220 camera. The network consists of an optical flow generation module and a structural pose recognition module. An overall schematic of the proposed method is shown in Figure 2. The workflow of network is as follows:

First, a 3D deformed mesh model is established, which is as same as the RC frame structure. Translation parameter $H$ of structure is considered, but rotation parameter $R$ of structure is ignored. Uniformly random $H$ decides the pose of structure. In this study, each control section is assumed to be rigid. Thus, all of the control section will not have the geometric shape changes. Spline interpolation is used to determine the translation of all the other common sections of the 3D structural model.

Second, the frames from videos are fed into FlowNet2 to generate optical flow with initial frame, respectively. These optical flows are denoised by a mask which is made manually.

Third, various deformed 3D models which can be rendered to generate rendered pictures are generated by different uniformly random parameters $H$. These rendered pictures are fed into Flow and denoised to generate optical flow in the way that is same as the second step. The training set consisted of inputs (denoised optical flow) and ground truths (parameters $H$) is used to train POFRN. The optical flow of video frames from the second step is inputted into the trained POFRN to output the predicted value of parameters $H$ of all the video frames.

Finally, the geometry of the structure is determined by the predicted value parameters $H$ from POFRN. The displacement of specified points of structure can be calculated.

The detail of Deformable 3D mesh model, dense optical flow, denoising process, and POFRNet are elaborated in the remainder of this section.

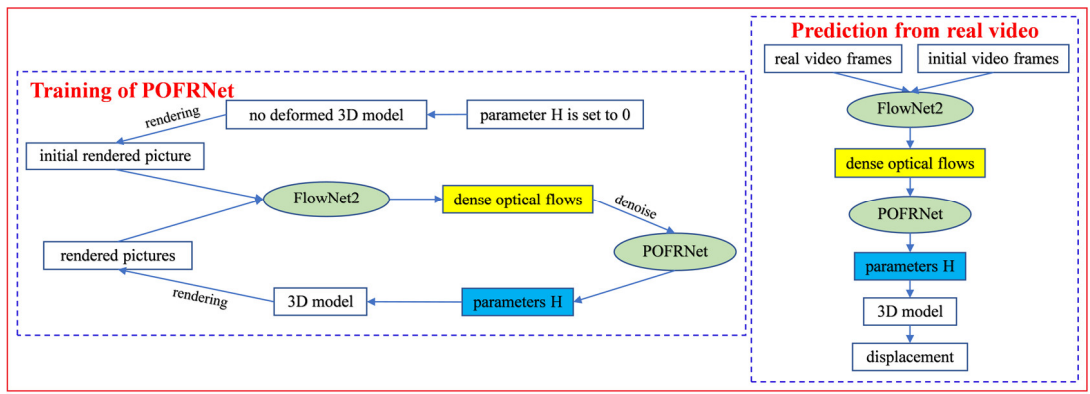

Fig. 2. Overall Schematic of the Proposed Method

### 3.2 Deformable 3D mesh model

On account of the dimension of RC frame structure is given. Therefore, a mesh model can be manually established in the 3D modeling software blender. In the model, we define 5 control section $S_c$ along the height direction, where is at the 0, 1.85, 3.35, 4.85, 6.50 meters position respectively as shown in Figure 3.

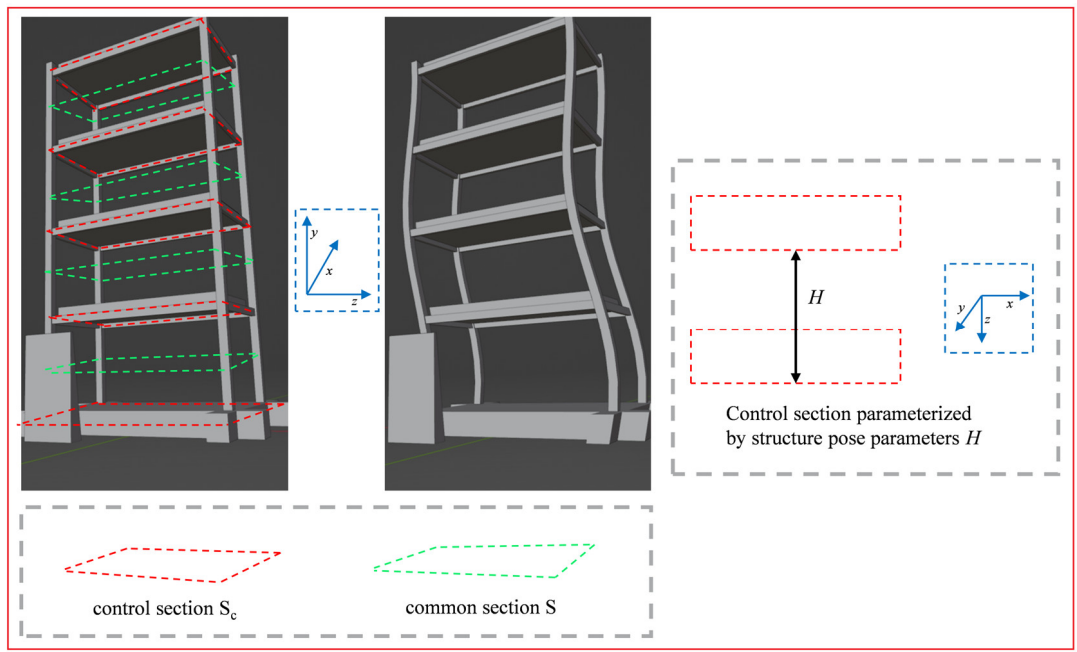

Fig. 3. 3D Mesh Model and Section

Assume that common sections and control sections perpendicular to the height direction are rigid and that these sections do not move along the length direction. Furthermore, suppose that torsional oscillation of the structure is ignored and only the translation offsets $H$ of the structure is taken into account. $H$ is normalized by the height of the structure. Any section along the height of the structure is calculated using cubic spline interpolation.

## 3.3 Dense optical flow

Dense optical flow represents that optical filed composed of displacement vector which each pixel in image *I* needs to move when image from *I* transform to *I'*. Taking the optical flow of two frames in Figure 4 as an example, the dense optical flow field of two frames of images can be obtained through optical flow calculation. The optical flow field can be converted to RGB images for visualization by color wheel conversion as shown in Figure 5, different hue represents the direction of optical flow, saturability represents the module of optical flow.

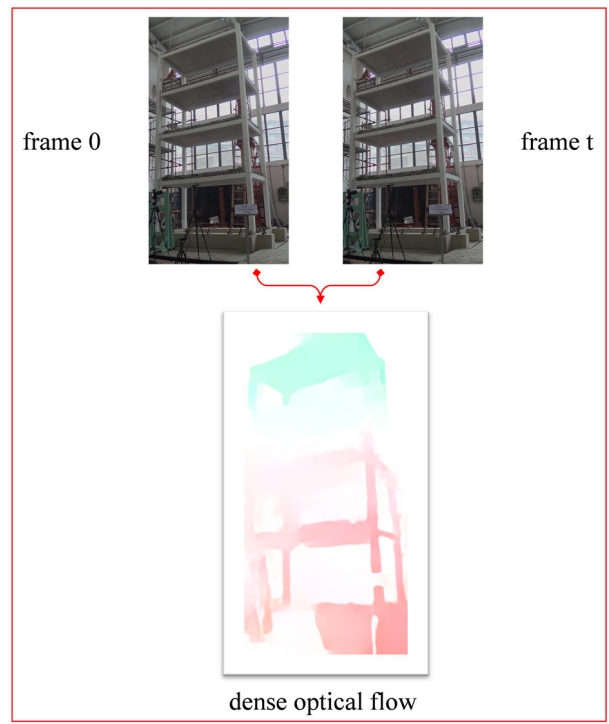

Fig. 4. The Schematic of Generation of Dense Optical Flow

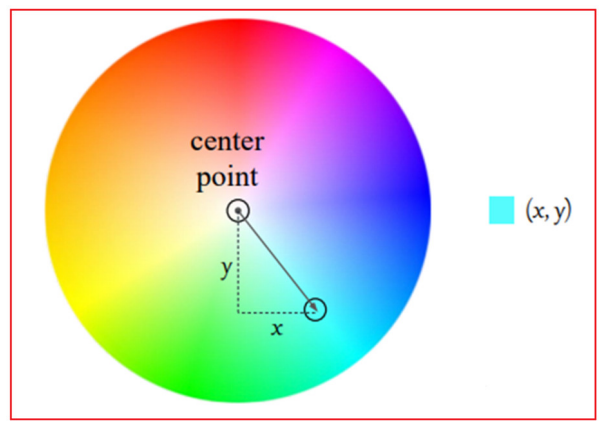

Fig. 5. Mapping between Dense Optical Flow Color
and Optical Flow Vector

## 3.4 FlowNet2 and Denoise

The FlowNet2 needs to take the video frames as inputs, then calculate the optical flow between every frame and initial frame through FlowNet2(initial frame can take the frame in which the camera is stable as well as the structure is not deformed state). On account of the existing of noise of the optical flow, the optical flow is denoised by a mask which are made manually by using software PS as shown in Figure 6, which means that the displacement vector of pixels at the structure position keep the same, but that of pixels at the non-structure position are adjusted to be zero.

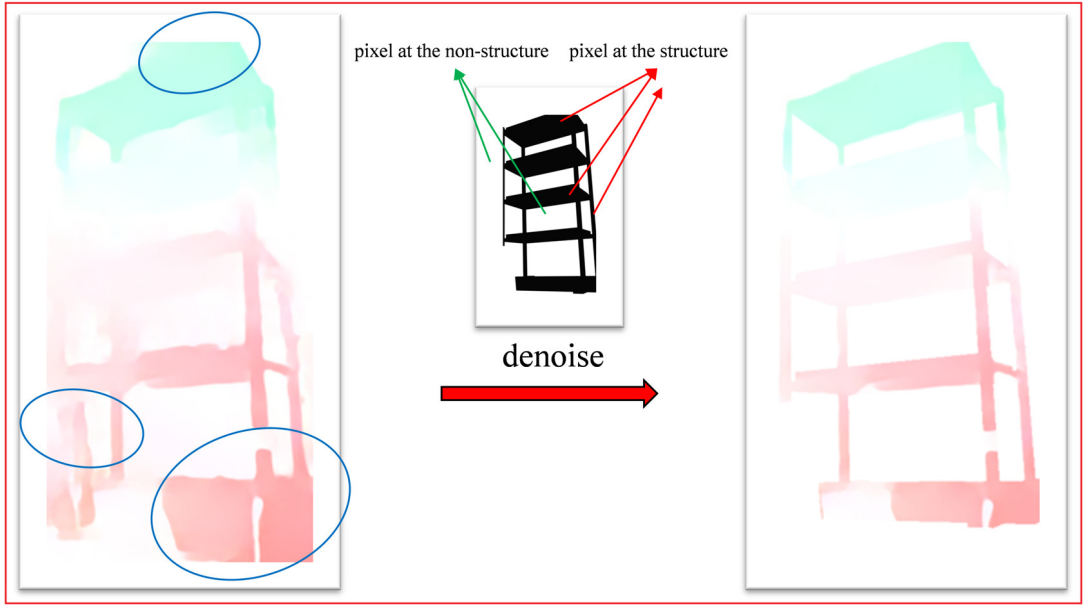

Fig. 6. The Schematic of Process of Denoising

### 3.5  ParaNet based on Optical Flow Removed Noise (POFRN)

The goal of POFRN is to obtain the pose parameters $H$ from the denoised optical flow:

$$[H] = POFRN(K) \tag{1}$$

Once this pose parameters are recognized, the model pose can be confirmed, then the displacement of any point on the structure can be calculated.

#### 3.5.1 Dataset

To train the POFRN, many training samples $K$ and their corresponding labels $H$ are required. For each deformed mesh model, the pose parameters $H$ consist of the following elements:

$$H = \{H_1, H_2, ..., H_N\} \tag{2}$$

where $N$ is the number of the control section.

To generate a dataset to train POFRN, many uniformly distributed $H$ are randomly generated first. The ranges of $H$ are hypermeters. Then, the generated $H$ are fed into 3D mesh model. 3D software Blender can be rendered these models to pictures with color only (without any light) as shown in Figure 7. These rendered pictures are fed into FlowNet2 with another rendered picture which $H$ is zero (that is no deformation model), respectively. The results of these pictures which are denoised by a mask is the denoised optical flow $K$, which corresponds to $H$.

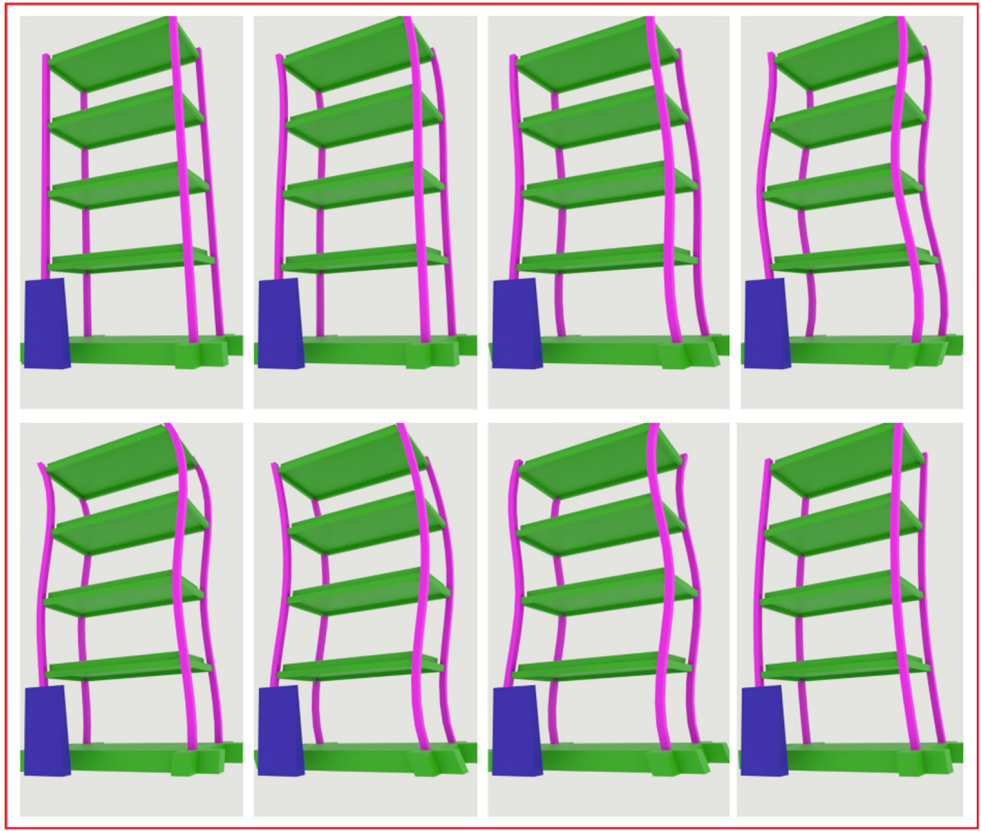

Fig. 7. Rendered Picture by 3D Software Blender

### 3.5.2 Network Structure

Inception V3 is modified and applied as the backbone of POFRN-Net as shown in Figure 8, which also exhibits the substructures of Inception A, Inception B, Inception C, Inception D, Inception E respectively.

In the POFRN-Net module, the mean square error (MSE) loss function is applied to estimate the difference between the ground truth value of the structure pose parameters $H$ and the predicted value from POFRN-Net:

$$\text{MSELoss}(H,\hat{H}) = \sum_{i=1}^{N} \frac{1}{N}\left(H_i - \hat{H}_i\right)^2 + \lambda_2 \|\boldsymbol{\omega}_{POFRN}\|_2^2 \qquad (3)$$

where $H$ are the predicted values and $\hat{H}$ are the ground truths. $\boldsymbol{\omega}_{POFRN}$ is the weight of the POFRN, and $\lambda_2 \|\boldsymbol{\omega}_{POFRN}\|_2^2$ is the L2 regularization term with a factor $\lambda_2$.

The output pose parameters $H$ of POFR-Net contains the structure pose information. It can be used to calculate the displacement of specified point, which is introduced in the next section.

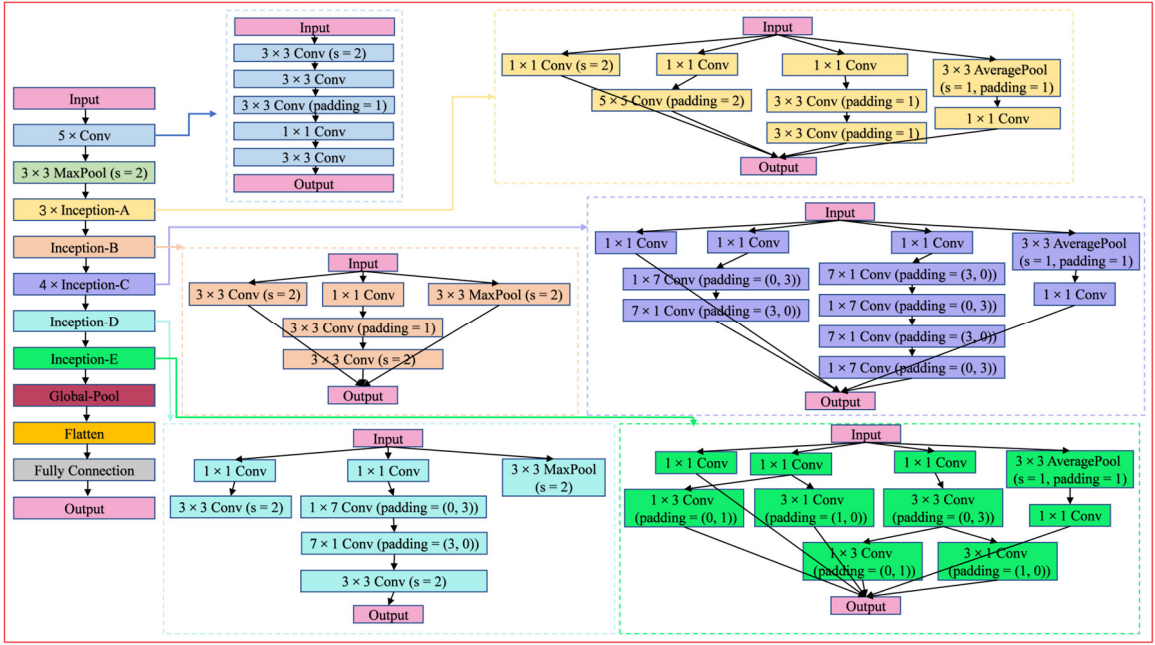

Fig. 8. Network structure of the POFR-Net

### 3.5.3 Displacement calculation

The coordinates ($x, y, z$) of a vertex $V$ on common section $S$ can be calculated from the predicted $H$, where y is the entity in the direction perpendicular to the section, z is the entity in the direction of $H$, and $x$ is the entity in the direction perpendicular to $y$ and $z$, which is shown as Figure 3. $H_S$ in the common section S is calculated by cubic spline interpolation function *CubSplItp*:

$$H_S = CubSplItp\,(\mathrm{y}_0, H) \qquad (4)$$

where $y_0$ is the entity of vertex $V$ on Section S in the model.

On account of the ignorant of rotation of structure, $x_0$ and $y_0$ is not expected to change. The coordinates of vertex $V$ can be updated as follows:

$$\begin{cases} x = x_0 \\ y = y_0 \\ z = z_0 + H_S \cdot h \end{cases} \qquad (5)$$

where h is the height of the structure. Therefore, the 3D displacement of any points of structure can be obtained from original coordinate and pose parameters $H$.

## 4 EXPERIMENT

As discussed previously, we created a 3D model of the RC frame structure with five control sections in Blender in this project. The heights of the control sections are 0.35, 1.85, 3.35, 4.85, and 6.55 meters, respectively.

In this case, the oscillation of the RC frame structure is along one lateral direction. Torsional oscillations can be neglected. Therefore, two sets of data with different uniform distribution ranges (the range can be seen as a hyper-parameter) of $H$ are generated to train the proposed net for the different video. The first set has $H$ uniformly distributed between -0.025 and 0.025, and the second set has $H$ between -0.004615 and 0.004615 for small displacement in video 1. Each set has 20,000 samples, with 20% used for testing. By randomly generating the different $H$, various rendering images with the 384×683 resolution in which structure have different pose are obtained. Extra zero pixels that make no difference along the height direction are added to transform the resolution from 384×683 to 384×703 which is a multiple of 64 required by flownet2.

All the rendered picture are separately paired with the initial rendered image, which are fed into FlowNet2 to generate corresponding dense optical flow which represents the position change information from the initial state of the model. These generated optical flows which are removed noise to prevent interference from moving objects except the model have one-to-one correspondence with $H$, which are used as the training set with the labels.

The hyperparameters of POFRN-Net are set as follows: the learning rate $\lambda$ is set to 0.01, and the batch size $b$ was set to 20. A learning rate decay strategy is also applied in the training process, and the decay rate is 0.94 for every epoch. The $L_2$ regulation factor $\lambda_2$ was set to 0.0001 to avoid overfitting. The total number of training epochs was 30, which also ensure that the training and test loss function converge. All the trainings are conducted on an NVIDIA RTX 3080 GPU. The curves of the training and testing loss under two kinds of H are shown in Figure 9.

Images are extracted frame by frame from the three videos. The original image size is $1920 \times 1080$. These images are rotated, removed distortion, resized, added zero pixels and the final size is $384 \times 704$ to be consistent with the rendered images. On account of the shake of monocular in the beginning of the video, the video frame in which the monocular dose not shake is chosen the initial frame. The $0^{th}$ frame, the $500^{th}$ frame and the $375^{th}$ frame is set to the initial frame of video1, video2 and video3, respectively. The $5375^{th}$ frame, the $1600^{th}$ frame and the $7721^{th}$ frame (last frame of video 3) is set to the last frame which is inputted into the FlowNet2 with corresponding initial frame, respectively. During the time before the initial frame and after last frame which is inputted into the FlowNet2, the RC frame structure is non-deformed, which the displacement of any point is zero. Using the same method as above, the optical flow with respect to the video can also be obtained.

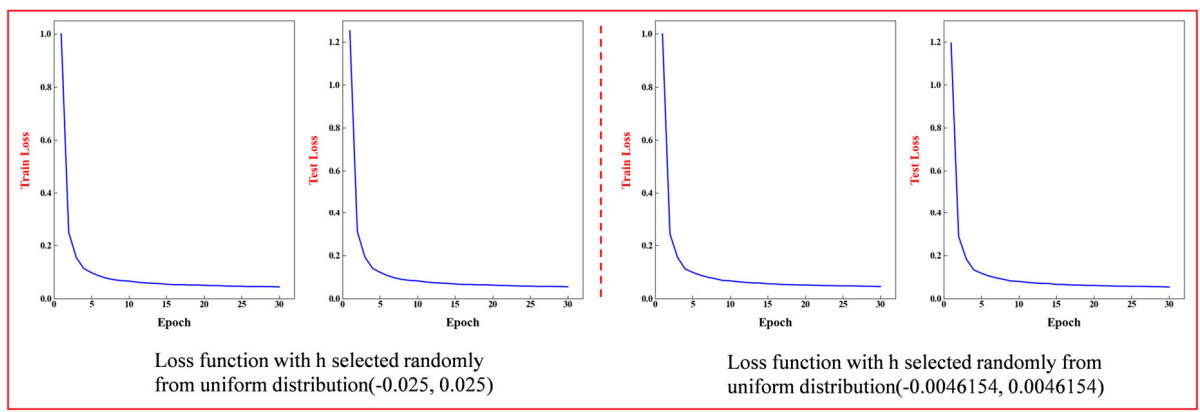

Fig. 9. Loss Function of POFRN-Net

## 5 RESULT AND DISCUSSION

According to the ground truth on the Floor 1 provided by Video 1, the comparison graph of the displacement curve of the true value and the predicted value can be drawn as shown in the figure 10. The results shows that the displacements are predicted with a certain degree of accuracy.

Figure 11, 12 and 13 illustrates the predicted displacement time-history curve and FFT results using the proposed method for the three videos, respectively. The time axis of these three predicted displacement curves are 0-215 seconds, 0-44 seconds and 0-293 seconds, which are correspond the 0-215 seconds, 20-64 seconds and 15-308 seconds in three videos, respectively.

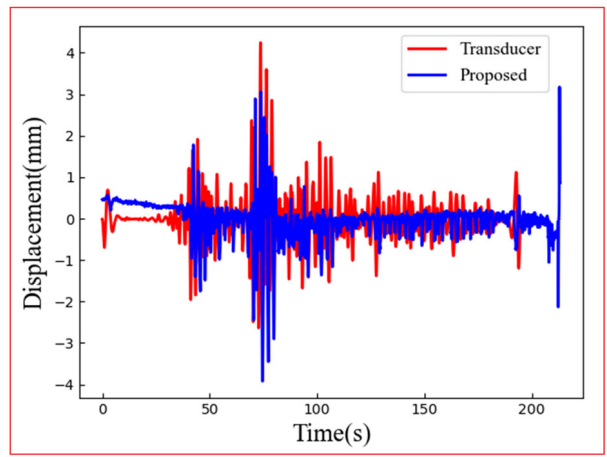

Fig. 10. Displacement of True Value and Predicted Value

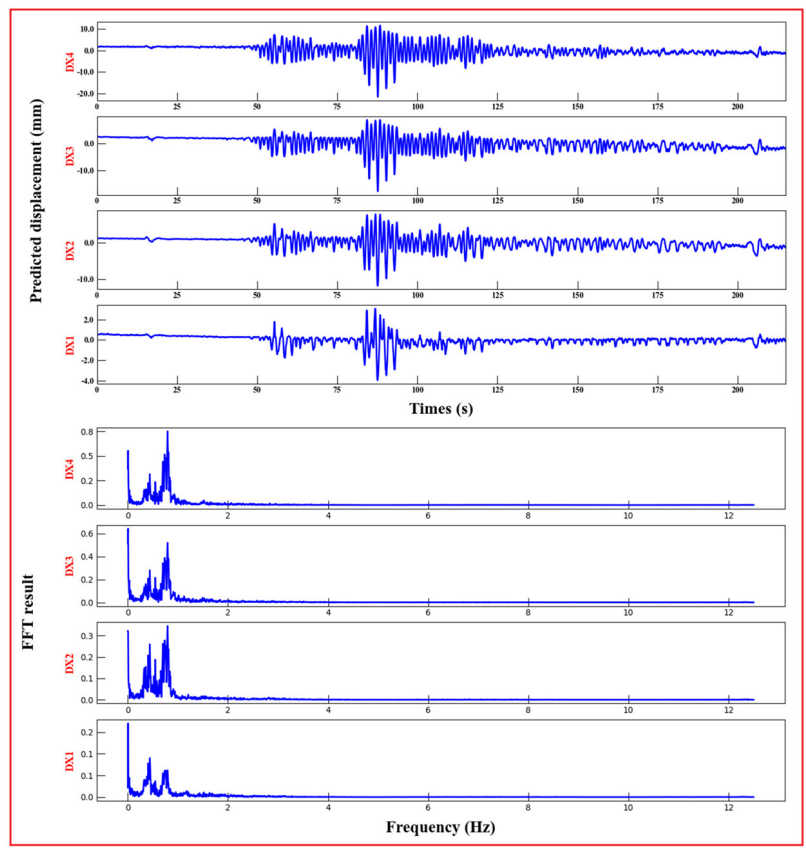

Fig. 11. Displacement and FFT Result for video 1

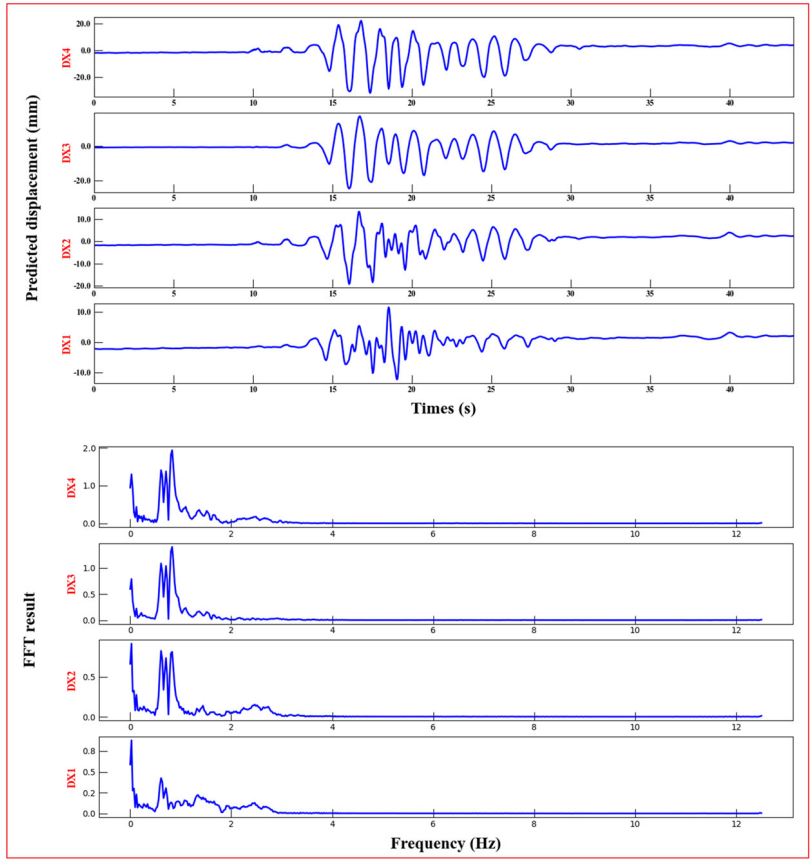

Fig. 12. Displacement and FFT Result for video 2

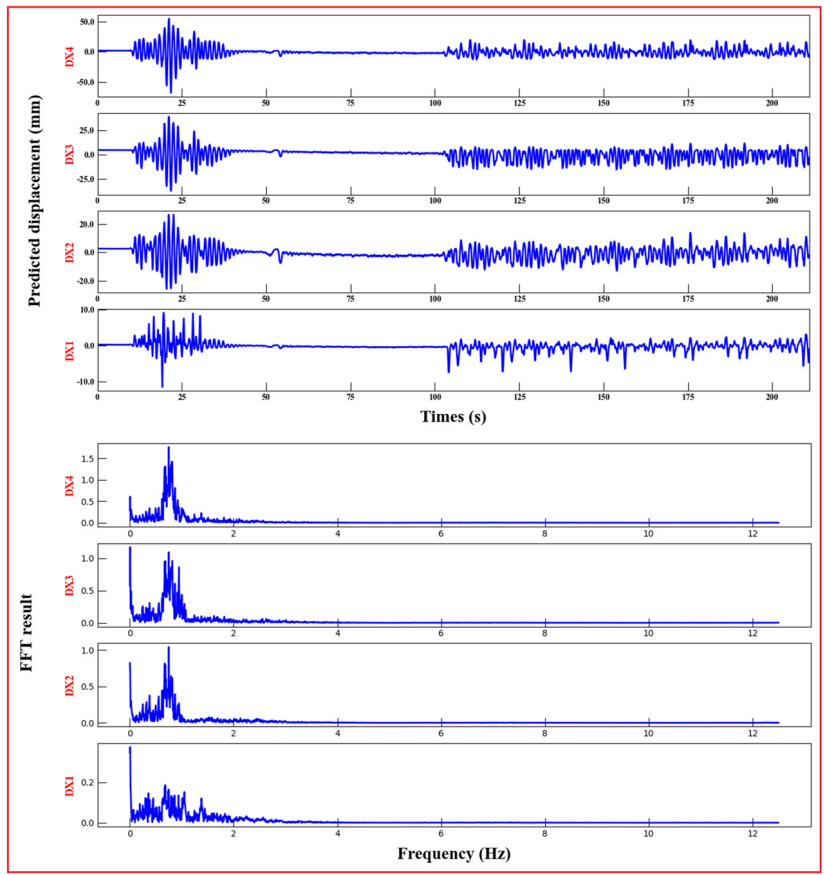

Fig. 13. Displacement and FFT Result for video 3

**Acknowledge**

The authors would like to thank Ph.D. Jin Zhao, Prof. Yang Xu from Harbin Institute of Technology for their valuable suggestions and patient help in the 3nd International Competition for Structure Healthy Monitor.